\pdfoutput=1

\documentclass[11pt]{article}
\usepackage{tikz}
\newcommand{\myboxans}[1]{\tikz[baseline=(MeNode.base)]{\node[rounded corners, fill=gray!20](MeNode){#1};}}

\newcommand{\mybox}[1]{\tikz[baseline=(MeNode.base)]{\node[rounded corners, fill=orange!20](MeNode){#1};}}

\usepackage[]{acl}
\usepackage{multicol}
\usepackage{soul}
\usepackage{multirow}
\usepackage{CJKutf8}
\usepackage{subfigure}
\usepackage{hyperref}
\usepackage{graphicx}
\usepackage{subfigure}
\usepackage{subcaption}
\usepackage{times}
\usepackage{latexsym}
\usepackage{booktabs}

\usepackage[T1]{fontenc}

\usepackage[utf8]{inputenc}

\usepackage{microtype}

\usepackage{inconsolata}
\title{Quantifying  Atomic Knowledge in  Self-Diagnosis \\
for Chinese Medical LLMs}

\author{
    Yaxin Fan$^{1}$, 
    Feng Jiang$^{2,3,4}$\thanks{\quad Corresponding author}, 
    Benyou Wang$^{2,3}$, 
    Peifeng Li$^{1}$\and 
    Haizhou Li$^{2,3}$
    \\
    $^1$School of Computer Science and Technology, Soochow University, Suzhou, China
    \\
    $^2$School of Data Science, The Chinese University of Hong Kong, Shenzhen, China
    \\
    $^3$Shenzhen Research Institute of Big Data, Shenzhen, China
    \\
    $^4$ University of Science and Technology of China, Hefei, China
    \\
    \texttt{yxfansuda@stu.suda.edu.cn}
    \\
    \texttt{\{jeffreyjiang,wangbenyou,haizhouli\}@cuhk.edu.cn}
    \\
    \texttt{pfli@suda.edu.cn}
}

\begin{document}
\begin{CJK}{UTF8}{gbsn}
\maketitle
\begin{abstract}
The booming development of medical large-scale language models (LLMs) enables users to complete preliminary medical consultations (self-diagnosis) in their daily lives. Recent evaluations of medical LLMs mainly focus on their ability to complete medical tasks, pass medical examinations, or obtain a favorable GPT-4 rating. There are still challenges in using them to provide directions for improving medical LLMs, including misalignment with practical use, lack of depth in exploration, and over-reliance on GPT-4. To address the above issues, we construct a fact-checking style Self-Diagnostic Atomic Knowledge (SDAK) benchmark. Through atomic knowledge that is close to real usage scenarios, it can more accurately, reliably, and fundamentally evaluate the memorization ability of medical LLMs for medical knowledge. The experimental results show that Chinese medical LLMs still have much room for improvement in self-diagnostic atomic knowledge.  We further explore different types of data commonly adopted for fine-tuning medical LLMs and find that distilled data enhances medical knowledge retention more effectively than real-world doctor-patient conversations.
\end{abstract}
\section{Introduction}

In the digital age, seeking health information from the Internet for self-diagnosis has become a common practice of patients~\citep{white2009experiences, 10.1093/jamia/ocz152, farnood2020mixed}. During the self-diagnosis, the searched health information can assist users in making necessary medical decisions, such as self-treatment or going to the hospital for professional treatment. With the development of generative models \citep{ouyang2022training, sun2021ernie, openai2023gpt4}, Large-scale Language Models (LLMs) hold the promise of revolutionizing the retrieval paradigm that seeks health suggestions via a search engine because they can provide more efficient suggestions through natural conversations.

\begin{figure}[!t]
	\centering  
        \includegraphics[width=1\linewidth]{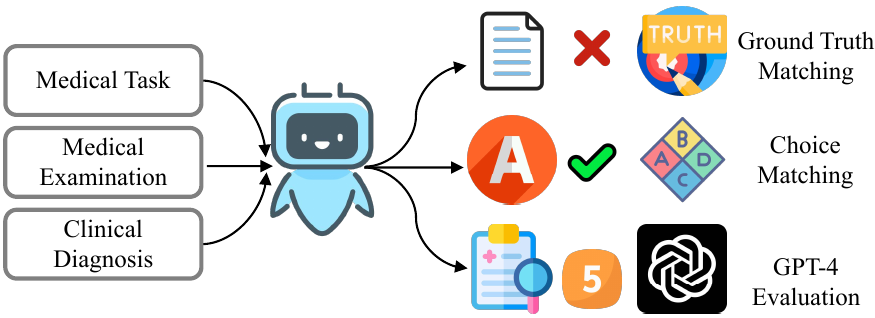}
        \caption{Widely used medical evaluation methods. The medical task mainly measures the ability of LLMs to complete the task, the medical examination explores the ability of LLMs to pass the examination, and the clinical diagnosis assesses the diagnosis ability of LLMs by using GPT-4 as the judgment.}
         \label{evaluationMethods}
          \vspace{-0.2cm}
    \end{figure}

To enhance the medical capabilities of open-source LLMs in Chinese, recent studies~\cite{wang2023huatuo, zhang2023huatuogpt, zhu2023ChatMed, yang2023zhongjing} attempt to fine-tune the foundation models on medical instruction or conversation data. As for the methods for evaluating their performance, the existing work is mainly divided into three categories:  medical NLP-related tasks~\cite{zhu2023promptcblue}, medical exams~\cite{umapathi2023medhalt, wang2023cmb}, and evaluating medical dialogue evaluations through GPT-4~\cite{zhang2023huatuogpt, yang2023zhongjing}, as shown in Figure~\ref{evaluationMethods}. 

\paragraph{Challenges.} Despite the progress of evaluation, there are still some challenges in using them to provide directions for improving medical LLMs: (1) \textit{Misalignment with practical use}. Most current Chinese medical LLMs are patient-centric, typically addressing questions related to medical consultations rather than complex and professional queries, such as "\textit{What should I take for a cold?}"(感冒应该吃什么药？). The results from these evaluations, such as NLP tasks or medical exams, do not match the actual needs of users. 
(2) \textit{Lack of depth in exploration}. Since most evaluations simply judge whether the model's responses to complex questions are correct or incorrect, it is challenging to determine whether errors stem from basic memorization failures or a lack of advanced reasoning abilities in LLMs~\cite{zheng2023does}.
(3) \textit{Over-reliance on GPT-4 for evaluation}. Evaluation by GPT-4 is not satisfactory because of its evaluation bias \cite{wang2023large} and its insufficient medical knowledge~(seen in Figure~\ref{atomicTypesperformance}). 


\paragraph{Solutions.} To address the above limitations, we propose a fact-checking style medical benchmark named the Self-Diagnostic Atomic Knowledge Benchmark (SDAK) to assess Chinese medical LLMs. Inspired by atomic fact-checking~\cite{chern2023factool}, 
we utilize atomic knowledge~\cite{min2023factscore}, an indivisible unit of information, for a more precise, reliable, and fundamental evaluation of an LLM's proficiency in medical knowledge (examples are shown in Table~\ref{ExampleOfClaims}).
To ensure that the evaluation is closer to the real usage scenario of medical LLMs, we adopt thematic analysis~\cite{braun2012thematic, zheng2023does} to extract the most commonly used atomic knowledge types from self-diagnostic queries. Then, we create atomic knowledge items for each type according to structured medical contents from public medical websites, each item consists of a pair of factual and counterfactual claims. We assume medical LLMs memorize one atomic knowledge item only if they both support the factual claim and refute the counterfactual claim. 
To reduce reliance on GPT-4, we designed two \textbf{necessary automation} indicators (instruction following rate, factual accuracy) and an \textbf{optional manual} metric (accuracy reliability). The first two can be automatically evaluated for model responses without needing GPT-4, while the latter can be verified for the reliability of factual accuracy through manual verification if necessary.

\paragraph{Results.}The experimental results show that: (a) the instruction following ability of most medical LLMs fine-tuned with domain data decreased to varying degrees compared to general LLMs, and the memorization ability of LLMs in the medical domain was not significantly improved; (b) the reliability of the answers after manual verifying mostly exceed 95\% indicating our metric is reliable to measure the memorization ability of LLMs.


\paragraph{Findings.} After an in-depth analysis of error types, knowledge types, and data sources for fine-tuning, we find the following three points: (1) Sycophancy is the primary cause of errors, whether it is in general or medical LLMs. (2) There is still a huge gap between the existing Chinese medical LLMs and GPT-4, although GPT-4 performs poorly in some more specialized medical knowledge. (3) Compared to real doctor-patient conversation data, distilled data from the advanced LLMs can better help open-source LLMs memorize more atomic knowledge. We believe that this is due to the fact that doctors are less likely to explain medical knowledge and diagnosis to patients in real doctor-patient conversations. The above insights could provide future research directions for the Chinese medical LLMs community. Our data, code, and model will be released in <\url{AnonymousURL}>.

\section{Related Work}

\begin{figure*}[ht!]
\setlength{\belowcaptionskip}{-0.2cm}
	\centering
	\includegraphics[width=16cm]{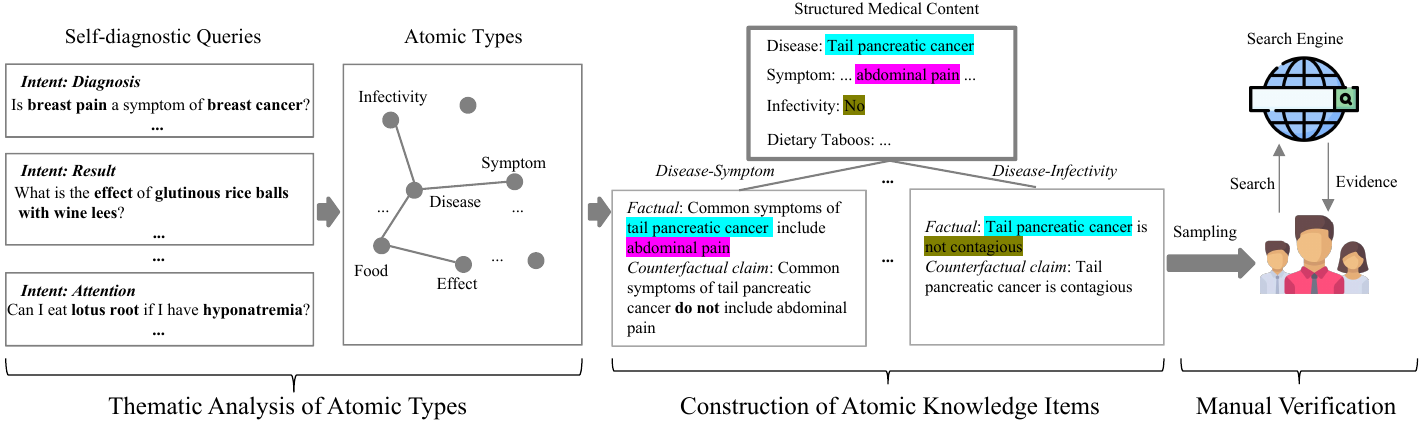}
	\caption{Construction process of self-diagnostic atomic knowledge benchmark.}
	\label{framework}
 \vspace{-0.2cm}
\end{figure*}
\subsection{Medical Evaluation Methods}

The existing efforts put into the evaluation of the medical abilities of LLMs are mainly divided into three types:  medical NLP-related tasks~\cite{zhu2023promptcblue}, medical exams~\cite{umapathi2023medhalt, wang2023cmb}, and conducting medical dialogue evaluations through GPT-4~\cite{zhang2023huatuogpt, yang2023zhongjing}. However, inconsistent scenarios, lack of depth
exploration, and insufficient medical ability of GPT-4 pose new challenges to evaluating medical LLMs in Chinese.
In this paper, we aim to address the above limitations and explore the memorization ability of LLMs in the self-diagnostic scenario.

\subsection{Fact-checking}
The fact-checking task \cite{thorne-etal-2018-fever, 10.1162/tacl_a_00454, wadden-etal-2020-fact, saakyan-etal-2021-covid, sarrouti-etal-2021-evidence-based, mohr-etal-2022-covert} aims to determine whether the claims are supported by the evidence provided, which has been an active area of research in NLP. 
Recently, some researchers \cite{min2023factscore,chern2023factool} have paid more attention to automatically evaluating the factuality of atomic knowledge contained in the long-form model-generated text. They utilized GPT-4 to automatically decompose atomic facts in complex texts and verify the overall factual accuracy.
However, this method is not applicable in the medical domain due to the insufficient mastery of medical knowledge in GPT-4, which cannot extract critical facts in user queries like extracting common sense.

\subsection{Chinese Medical LLMs}

To enhance the medical capability of open-source LLMs \cite{du2022glm, touvron2023llama1, touvron2023llama2, baichuan2023baichuan2}, previous work has attempted to adopt real-world medical data or the mixture of real-world and distilled/semi-distilled from ChatGPT conversations \cite{wang-etal-2023-self-instruct, wang2023med} for fine-tuning. The former \cite{MedicalGPT, wang2023huatuo, wang2023med} mainly learn the medical capabilities of doctors from doctor-patient conversations, while the latter \cite{zhu2023ChatMed, yang2023zhongjing, zhang2023huatuogpt} further additionally added distilled conversations from advanced LLMs such as ChatGPT. Despite there being much progress in medical LLMs in Chinese, how to better evaluate their performance is still an area that needs to be studied, such as the extent of self-diagnostic medical knowledge stored in these LLMs.

\section{Construction of Self-diagnostic Atomic Knowledge Benchmark}

\textbf{Motivation.} Despite the robust growth of Chinese medical LLMs, various evaluations for them have yet to be significantly helpful in improving them. On the one hand, some existing evaluations focus on medical NLP tasks~\cite{zhu2023promptcblue} or medical exams~\cite{umapathi2023medhalt, wang2023cmb}, which do not align with real usage scenarios (self-diagnosis). Moreover, due to the complexity of the testing questions, it is challenging to determine whether the model's errors stem from issues in memory, or reasoning, which are crucial for improving LLMs' performance. On the other hand, some efforts~\cite{zhang2023huatuogpt, yang2023zhongjing} have attempted to use GPT-4 in a conversational format for evaluation. However, due to GPT-4's inherent evaluation biases, its imperfect grasp of medical knowledge, and limitations in accessibility, this method is also not suitable. Therefore, inspired by atomic fact-checking evaluation studies, we build a fact-checking style medical benchmark named the Self-Diagnostic Atomic Knowledge Benchmark (SDAK) to more accurately, reliably, and fundamentally evaluate the memorization ability of medical LLMs for medical knowledge, as shown in Figure~\ref{framework}.


\subsection{Thematic Analysis of Atomic Types}

To obtain the most common types of atomic knowledge for queries of real users in the self-diagnostic scenario, we select the KUAKE-QIC~\cite{zhang-etal-2022-cblue} dataset as the source data. It mainly contains user queries from search engines with ten intent types, and examples are shown in Appendix~\ref{examplequeryappendix}.


Then, we conducted thematic analysis~\cite{braun2012thematic, zheng2023does} of 200 samples randomly selected from each intent type in KUAKE-QIC to identify the atomic knowledge types. 
Specifically, we first conduct the induction by initiating the preliminary type of atomic knowledge for each selected sample, where we mainly focus on medical-related knowledge, specializing in Disease-Symptom, Medicine-Effect, etc. Then, we deduce the most common type of atomic knowledge by aggregating the type into a broader atomic type if more samples fall into this type. 
Take the query with \emph{Diagnosis} intent in Figure~\ref{framework} as an example. Since both \emph{breast pain} and \emph{ breast cancer} in this query are the symptom and disease, respectively, the atomic type involved in this query is Disease-Symptom. 

Table~\ref{QueryTypes} shows the atomic types and percentages contained in the queries with various intents we constructed. We find that over 80\% of queries in each intent fall into different atomic types we deduced, indicating that atomic knowledge is a more fine-grained basic unit.  Besides, the queries with different intents tend to involve the same type of atomic knowledge, e.g., queries with both \emph{Diagnosis} and \emph{Cause} intents involve the same atomic type of Disease-Symptom, which demonstrates the necessity and efficiency of evaluating LLMs in terms of atomic knowledge. After removing the non-objective intent related to specific user locations, such as \emph{Price} and \emph{Advice}, we collect 17 most common types of atomic knowledge from real-world self-diagnostic queries, as shown in Table~\ref{QueryTypes}.

\begin{table}[t] \footnotesize
\centering
\setlength{\tabcolsep}{0pt}
\begin{tabular}{llr} 
\hline
\textbf{Intent}               & \textbf{Atomic Type} & \textbf{Percentage} \\ 
\hline
\multirow{2}{*}{\begin{tabular}[l]{@{}c@{}}  Diagnosis \end{tabular} }& Disease-Symptom  & 81\%       \\
                               &Disease-Examination   & 10\%       \\ \hline
\multirow{2}{*}{\begin{tabular}[l]{@{}c@{}}  Cause \end{tabular} }    & Disease-Cause  & 64\%       \\
                                 & Disease-Symptom  & 25\%       \\ \hline
\multirow{2}{*}{\begin{tabular}[l]{@{}c@{}} Method \end{tabular} }  & Disease-Medicine  & 55\%       \\
                                 &Disease-Method   & 34\%       \\ \hline
\multirow{3}{*}{\begin{tabular}[l]{@{}c@{}} Advice \end{tabular} }   & \st{Disease-Hospital}  & 80\%       \\
                                 &Disease-Department   & 8\%       \\
                                  &Disease-Examination   & 11\%       \\ \hline
\multirow{2}{*}{\begin{tabular}[l]{@{}c@{}} Metric\_explain \end{tabular}  }  & Examination-Range  & 63\%       \\
                                 &Metric-Effect   & 37\%       \\ \hline
\multirow{3}{*}{\begin{tabular}[l]{@{}c@{}} Disease\_express \end{tabular} }  & Disease-Symptom  & 62\%   \\
                                 &Disease-Infectivity   & 15\%       \\
                                  &Diseases-Complication   & 15\%       \\ \hline
\multirow{4}{*}{\begin{tabular}[l]{@{}c@{}} Result \end{tabular} }  & Disease-Symptom  & 36\%       \\
                                 &Western Medicine-SideEffect   & 14\%       \\
                                  &Chinese Medicine-SideEffect   & 19\%       \\
                                  &Food-Effect   & 17\%       \\ \hline
\multirow{2}{*}{\begin{tabular}[l]{@{}c@{}} Attention \end{tabular} }                  & Disease-Food  & 59\%       \\
                                 &Disease-Prevention   & 21\%       \\ \hline
\multirow{3}{*}{\begin{tabular}[l]{@{}c@{}} Effect \end{tabular} }                  & Western Medicine-Effect & 20\%       \\
        & Chinese Medicine-Effect & 27\%       \\
       &Food-Effect  & 44\%       \\ \hline
\begin{tabular}[l]{@{}c@{}} Price \end{tabular}                      & \st{Treatment-Price}              & 97\%         \\ 
\hline
\end{tabular}
\caption{Major types and percentages of atomic knowledge contained in each intent of self-diagnostic queries.}
\label{QueryTypes}
\vspace{-0.3cm}
\end{table}

\subsection{Construction of Atomic Knowledge Items}
After obtaining the most common atomic types, we construct pairs of factual and counterfactual claims for each atomic type to convey atomic knowledge items. 
To avoid data contamination, we do not construct atomic claims based on existing Chinese medical knowledge graphs, e.g., CMeKG~\cite{Odmaa2019} has been utilized by some Chinese medical LLMs~\cite{wang2023huatuo,wang2023med}. Instead, we manually build atomic knowledge items according to the structured medical content from the public medical websites\footnote{https://www.xiaohe.cn/medical \\and https://www.120ask.com/disease} for the following two reasons. On the one hand, the medical content from these websites is reliable because it is edited and verified by professional medical teams. On the other hand, these websites are also the main source of medical knowledge for self-diagnostic queries.

\begin{table}[t!]
\centering
\setlength{\tabcolsep}{3pt}
\resizebox{\linewidth}{!}{
\begin{tabular}{ll} 
\hline
\textbf{Atomic Type}  & \textbf{Example of factual (Counterfactual) atomic claim}  \\ 
\hline
Disease-Symptom    & \begin{tabular}[l]{@{}l@{}} Common symptoms of tail pancreatic cancer (do not) include  abdominal pain \\ 胰尾癌的常见症状（不）包括腹痛\end{tabular}   \\ 
Disease-Infectivity    & \begin{tabular}[l]{@{}l@{}}Laryngeal cysts are (not) contagious \\ 喉囊肿（不）具有传染性     \end{tabular}   \\
Disease-Department   & \begin{tabular}[l]{@{}l@{}}  Common departments for Psoriatic A (do not) include dermatology  \\银屑病甲常挂的科室（不）包括皮肤科    \end{tabular}     \\
Disease-Method   & \begin{tabular}[l]{@{}l@{}}  Common treatments for prolactinomas (do not) include radiation therapy \\ 催乳素瘤常见治疗方法（不）包括放射治疗    \end{tabular}     \\
\begin{tabular}[c]{@{}c@{}}......\end{tabular}    & \begin{tabular}[l]{@{}c@{}}......\end{tabular}      \\
Disease-Medicine  & \begin{tabular}[l]{@{}l@{}}  Common medications for stomatitis (do not) include metformin \\ 口腔炎的常用药物包括（不包括）甲氰咪胍    \end{tabular}       \\
\hline
\end{tabular}
}
\caption{Example of each type of atomic knowledge. The complete table is shown in Appendix~\ref{statisticOfSDAK}.}
\label{ExampleOfClaims}
\end{table}


As shown in Figure~\ref{framework}, we first extract the atomic knowledge from the structured medical content according to the atomic types we build. For example, we extract the disease \emph{Tail pancreatic cancer (尾胰癌)} and symptom \emph{abdominal pain (腹痛)} for the Disease-Symptom atomic type. Then, we heuristically construct a factual claim in the form of implication relation, as shown in Table~\ref{ExampleOfClaims}. 


Given that LLMs may exhibit a sycophantic bias \cite{wei2023simple, du2023calla}, e.g., it always supports the user's claims, it is unreliable to explore the amount of self-diagnostic knowledge stored in LLMs' memory merely by whether or not LLMs supports factual claims. To avoid this, we propose using contrastive evaluation based on constructing a counterfactual claim for each factual claim by converting the \emph{implication} into a \emph{non-implication} relation, as shown in Table~\ref{ExampleOfClaims}. LLMs are considered to possess one atomic knowledge item only if they both support the factual claim and refute the counterfactual claim.  For each atomic type, we randomly selected at most  1,000 structured medical content to build atomic knowledge items and the statistics are shown in Appendix~\ref{statisticOfSDAK}.

\begin{figure*}[ht!]
\setlength{\belowcaptionskip}{-0.2cm}
	\centering
	\includegraphics[width=14cm]{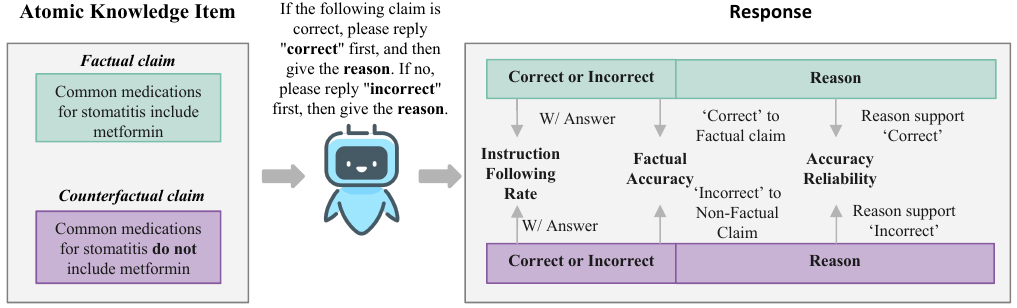}
	\caption{Process of the fact-checking style evaluation method.}
	\label{EvaluationProcess}
 \vspace{-0.2cm}
\end{figure*}

\subsection{Manual Verification}
To verify the reliability of atomic claims, we conducted the manual verification based on the evidence retrieved through a search engine.
We first randomly selected 50 factual claims for each atomic type. 
Then, we verify the correction of the claims. Follow the previous work \cite{chern2023factool} and retrieve evidence by feeding factual claims into a search engine~\footnote{The search engine we adopted is \href{https://www.baidu.com}{Baidu}, which is one of the most popular Chinese search engines.}. The top 10 items retrieved by the search engine as evidence and manually judge whether the evidence supports the factual claims. 

\begin{table}[t!]\scriptsize
\centering
\begin{tabular}{lccc}\hline
\multirow{2}{*}{\textbf{Atomic Type}} & \multicolumn{3}{c}{\textbf{Number}} \\
                             & \textbf{Support}  & \textbf{Neural} & \textbf{Refute} \\ \hline
Metric-Effect               & 43       & 6      & 1      \\
Disease-Infectivity            & 42       & 5      & 3      \\
Disease-Department           & 48       & 0      & 2      \\
Disease-Method               & 45       & 5      & 0      \\
Disease-Cause                & 46       & 4      & 0      \\
Chinese Medicine-Effect      & 48       & 1      & 1      \\ 
Chinese Medicine-SideEffect  & 46       & 1      & 3      \\
Western Medicine-Effect      & 50       & 0      & 0      \\
Western Medicine-SideEffect  & 44       & 3      & 3      \\
Food-Effect                  & 43       & 5      & 2      \\
Disease-Examination          & 45       & 0      & 5      \\
Disease-Prevention           & 33       & 11     & 6      \\
Diseases-Complication        & 42       & 7      & 1      \\
Disease-Symptom              & 48       & 2      & 0      \\
Examination-Range            & 32       & 12     & 6      \\
Disease-Food                 & 47       & 3      & 0      \\
Disease-Medicine             & 46       & 1      & 3      \\
\hline
Total                        & 748      & 66     & 36     \\
Percentage                   & 88.00\%   & 7.76\%  & 4.24\% \\
\hline
\end{tabular}
\caption{Manual verification of atomic knowledge items.}
\label{manuallyEvaluationofAtomicKnow}
\vspace{-0.2cm}
\end{table}

Table~\ref{manuallyEvaluationofAtomicKnow} shows the results of manual verification, where \emph{Support}, \emph{Neural}, and \emph{Refute} indicate that evidence supports claims, insufficient evidence, and evidence refutes claims, respectively. 88\% of claims can be fully supported by the evidence~\footnote{We also asked a professional doctor to verify 170 factual claims (each atomic type contains 10) and found there are 87\% of claims that can be supported.} and only 4\% are refuted, which shows the reliability of the atomic claims we constructed.
In addition, the reliability of about 8\% of factual claims cannot be verified due to insufficient evidence. We attribute it to the fact that these pieces of atomic knowledge are relatively low-frequency, leading to search engines failing to retrieve the related evidence.

\section{Experiments}

\subsection{General and Medical LLMs for Evaluation}

We select the following popular general LLMs and Chinese medical LLMs for evaluation on our SDAK. In addition to the closed-sourced ChatGPT and GPT-4~\cite{openai2023gpt4} models, we select representative open-source Chinese LLMs such as Baichuan2~\cite{baichuan2023baichuan2}, Qwen~\cite{bai2023qwen}, and ChatGLM2~\cite{du2022glm} for evaluation. As for the Chinese medical LLMs, we select two types of models: 

\textbf{Fine-tuned merely on real-world data}: BenTsao \cite{wang2023huatuo}, ChatGLM-Med \cite{wang2023med}, MedicalGPT~\cite{MedicalGPT}. 

\textbf{Fine-tuned on mixed data}: Chatmed-Consult \cite{zhu2023ChatMed}, HuatuoGPT \cite{zhang2023huatuogpt}, and Zhongjing \cite{yang2023zhongjing}.

We conducted the experiment in zero-shot and few-shot settings, and Appendix~\ref{Few-shot Experiments} introduces the details about few-shot setting. Appendix~\ref{Hyper-parameters} introduces the hyperparameter settings of each model.

\subsection{Fact-checking Style Evaluation Method}
\label{evaluationprompt}
To comprehensively evaluate the performance of LLM on the SDAK benchmark, we propose the fact-checking style evaluation method, as shown in Figure~\ref{EvaluationProcess}.

\subsubsection{Evaluation Prompt}
\label{section4.2.1}
Firstly, for a pair of claims for each atomic knowledge in SDAK, we designed an appropriate prompt to instruct LLMs to output as we specified to evaluate the performance of LLMs. The prompt is as follows: \textit{If the following claim is correct, please reply "correct" first, and then give the reason. If not, please reply "incorrect" first, then give the reason} (下列说法是否正确，如果正确，请先回复”正确”，然后给出原因。如果错误，请先回复“错误”，然后给出原因。).  The prompt specifies two parts of the output given by LLMs: the \textbf{answer} and the \textbf{reason}. 
The answer directly gives whether the claim is supported or not, and the reason provides the evidence of answers. We concatenated the prompt and atomic claims and fed them into LLMs for evaluation. Refer to Appendix~\ref{differentprompts} for the exploration of different prompts.

\begin{table*}[ht!] 
\footnotesize
\centering
\setlength{\tabcolsep}{2pt}
\begin{tabular}{cclrccrcc}
\hline
\multirow{2}{*}{\textbf{Domain}}  & \multirow{2}{*}{\textbf{Data}}  & \multirow{2}{*}{\textbf{LLMs}} & \multicolumn{3}{c}{\textbf{Zero-shot}}    & \multicolumn{3}{c}{\textbf{Few-shot}}     \\
                         &                        &                       & \textbf{IFR(\%)} & \textbf{FactAcc(\%)} & \textbf{AccR(\%)} & \textbf{IFR(\%)} & \textbf{FactAcc(\%)} & \textbf{AccR(\%)} \\
\multirow{7}{*}{General} & \multirow{7}{*}{-}     & GPT-4                & $99.96_{±0.00}$         & $65.42_{±0.60}$ & 100         & $100_{±0.00}$         &$72.61_{±0.33}$  & 100        \\
                         &                        & Qwen-14b-Chat         & $100_{±0.00}$         & $57.29_{±0.03}$         &    98 & $100_{±0.00}$         & $67.34_{±0.56}$         &    100          \\
                         &                        & ChatGPT               &  $99.97_{±0.00}$              &  $51.72_{±0.40}$                   &  97 &  $100_{±0.00}$              &   $56.93_{±0.79}$                 &  99         \\
                         &                        & Qwen-7b-Chat           & $100_{±0.00}$           & $43.68_{±0.10}$         &  98       & $100_{±0.00}$           & $56.74_{±0.30}$          &  100               \\
                         &                        & Baichuan2-13b-Chat    & $99.71_{±0.00}$       &$42.01_{±0.05}$          & 96      &$100_{±0.00}$       &$52.09_{±0.45}$        & 99              \\
                         &                        & ChatGLM2             & $99.84_{±0.01}$    & $37.90_{±0.04}$         & 97            & $100_{±0.00}$    & $47.17_{±2.13}$         & 100                     \\
                         &                        & Baichua2-7b-Chat     & $99.89_{±0.01}$   & $16.14_{±0.09}$         & 95      & $100_{±0.00}$   & $35.15_{±2.52}$         & 98              \\ \hline 
\multirow{6}{*}{Medical} & \multirow{3}{*}{Mixed} & Zhongjing            & $90.22_{±0.17}$         & $24.78_{±0.10}$              & 97 & $93.59_{±0.80}$         & $29.56_{±3.65}$              & 100         \\
                         &                        & Chatmed-Consult       & $85.10_{±0.14}$    & $24.50_{±0.34}$         & 98 & $95.32_{±2.31}$    & $27.15_{±2.91} $        & 99          \\
                         &                        & HuatuoGPT          & $99.73_{±0.00}$  & $16.15_{±0.01}$         & 98         & $99.90_{±0.62}$  & $26.63_{1.52}$         & 100             \\ \cline{2-9}         
                         & \multirow{3}{*}{Real}  & MedicalGPT            & $76.04_{±0.50}$    & $7.86_{±0.50}$           & 100    & $85.12_{±0.74}$     & $11.37_{±2.09}$           & 100              \\
                         &                        & ChatGLM-Med          & $94.91_{±0.07}$    & $7.46_{±0.15}$         & 75    & $97.79_{±0.68}$    & $9.41_{±0.89}$         & 93            \\
                         &                        & BenTsao                     & $84.43_{±0.06}$    & $3.35_{±0.07}$          & 70   & $89.37_{±3.20}$    & $7.26_{±3.49}$          & 96      \\ \hline     
\end{tabular}
\caption{The performance of general and medical LLMs on self-diagnostic atomic knowledge. The subscript represents the standard deviation after three experiments.}
\label{performance}
\vspace{-0.4cm}
\end{table*}

\subsubsection{Evaluation Metrics}

To evaluate the performance of LLMs in processing atomic knowledge, we developed two \textbf{necessary automatic} metrics: Instruction Following Rate (IFR) and Factual Accuracy (FactAcc), and an \textbf{optional manual} metric: Accuracy Reliability (AccR). These metrics collectively assess an LLM's ability to process and respond to medical information accurately and reliably. \textbf{Instruction Following Rate (IFR)} assesses whether LLMs can adhere to the given instructions. An LLM is considered to follow instructions if it provides answers (be it correct or incorrect) to both factual and counterfactual atomic claims at the start of its response. \textbf{Factual Accuracy (FactAcc)} measures the abilities of LLMs on self-diagnostic atomic knowledge. LLMs are considered to memorize the atomic knowledge if they give the answer \emph{'correct'} to the factual claim and \emph{'incorrect'} to the counterfactual claim of an item. \textbf{Accuracy Reliability (AccR)} evaluates the reliability of factual accuracy. We randomly selected 100 atomic knowledge items and manually checked the model's responses. If the reason given by LLMs can support the answer '\emph{correct}' to a factual claim and the answer '\emph{incorrect}' to a counterfactual claim, we believe that the answers given by LLMs are reliable.

\subsection{Evaluation Results}

The performance of each model on SDAK is shown in Table~\ref{performance}.  A key finding is that while general LLMs maintain an instruction-following rate above 99\% in zero-shot and few-shot settings, most medical LLMs show a 5\%-15\% decline in zero-shot setting. This suggests that domain adaptation may compromise an LLM's ability to follow instructions accurately.

In terms of factual accuracy (FactAcc) in the zero-shot setting, GPT-4 unsurprisingly achieves the best performance of 65.42\% among all LLMs. Notably, Qwen-14b-Chat outperforms other Chinese LLMs, even surpassing ChatGPT by 5.57\%. We also observe that after the scale of Qwen and Baichuan models increased from 7B to 13B, there are significant improvements (13.61\% and 25.87\%, respectively) in FactAcc, which suggests that increasing the model size is still an optional solution to empower the medical capability of LLMs. 

Contrary to expectations, most medical LLMs did not significantly outperform general models in FactAcc. Where Zhongjing, Chatmed-Consult, and HuatuoGPT surpass the Baichuan2-7b-chat, and their best performance (Zhongjing) in the FactAcc only reaches 24.78\% in the zero-shot setting. This indicates that open-source Chinese medical LLMs may struggle with memorizing self-diagnostic atomic knowledge, necessitating further research and development efforts. The significant differences in medical models with different training data prompted us to conduct an in-depth analysis of the impact of different training data, as shown in Section~\ref{differenttypeofdata}.




\begin{table*}[!t] \footnotesize
\setlength{\tabcolsep}{2pt}
\centering
\begin{tabular}{clcccc}
\toprule
\multirow{2}{*}{\textbf{Domain}}         & \multirow{2}{*}{\textbf{LLMs}}     & \multicolumn{4}{c}{\textbf{Error Type}}\\ &        & \textbf{NotFollow } & \textbf{Sycophancy} & \textbf{Safety} & \textbf{Misinterpretation}  \\
\midrule
\multirow{5}{*}{General}  & GPT4          &0  & 68                    & 26                        & 6                      \\
& Qwen-14b-Chat                   &0 & 68             &  24             & 8          \\
                          & ChatGPT     &0 & 79        & 17             & 4                          \\
                          & Qwen-7b-Chat   &0     &  74        &  20     & 6       \\
                          & Baichuan2-13b-Chat  &0  & 72       & 24    & 4            \\
                          & ChatGLM2-6b          &0  & 70     & 25    & 5                  \\
                          & Baichuan2-7b-Chat   &0  & 74     & 21        & 5             \\ \hline
\multirow{7}{*}{Medical} & Chatmed-Consult   & 20   & 48        & 18     & 14     \\
                          & Zhongjing            &8  &  62         &  18      &  12         \\
                          & HuatuoGPT        & 0  & 64         & 22     & 14    \\
                          & MedicalGPT       & 23 & 62        & 2          & 13    \\
                          & ChatGLM-med      & 5  & 54          & 1       & 40           \\
                          & BenTsao         & 5   & 90             & 5    & 0       \\
                         \bottomrule                                             
\end{tabular}
\caption{Error analysis of LLMs on atomic knowledge. }
\label{errortypes}
\end{table*}
In addition, in the few-shot setting, the performance of most models on all metrics is improved significantly, indicating that in-context learning can effectively improve the abilities of models on instruction-following and self-diagnostic atomic knowledge.

Finally, after manually checking the answers provided by various models, we find that both the general LLMs and the medical LLMs can provide a good basis for the answers, with most of them achieving over 95\% performance in Accuracy Reliability (AccR). It also proves that FactAcc can reliably reflect the LLMs' memorization ability of self-diagnostic atomic knowledge.


\section{Analysis}
The analysis is conducted in the zero-shot setting.

\subsection{Error Analysis on Atomic Knowledge}
\label{sectionerrortypes}

We conducted a detailed analysis of errors to gain insights into the challenges LLMs face in memorizing medical atomic knowledge. We randomly selected 100 atomic knowledge items where various models provided incorrect responses, as shown in Table~\ref{errortypes}. This analysis revealed four primary error categories: \textbf{NotFollow}, where LLMs either evade directly answering ('correct' or 'incorrect') or provide irrelevant information; \textbf{Sycophancy}, characterized by LLMs indiscriminately supporting both factual and counterfactual claims, distinct from mere bias or agreeability; \textbf{Safety}, LLMs argue that claims are not strictly expressed and provide a more cautious answer; and \textbf{Misinterpretation}, where LLMs erroneously treat counterfactual claims as factual. Appendix ~\ref{DetailsofErrorTypes} shows the examples of each type.

\begin{figure}[!t]
	\centering  
        \includegraphics[width=0.95\linewidth]{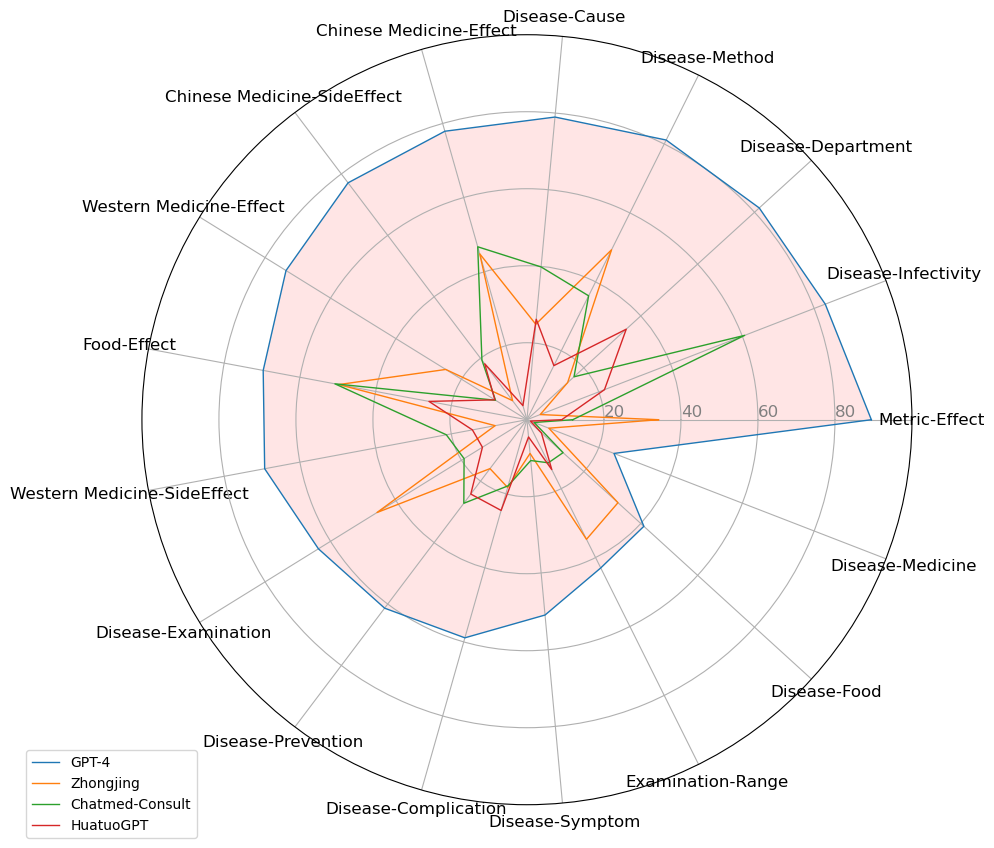}
        \caption{ Performance of representative LLMs on various types of atomic knowledge. }
         \label{atomicTypesperformance}
          \vspace{-0.4cm}
    \end{figure}
Table~\ref{errortypes} shows that the proportion of 'NotFollow' responses aligns with the Instruction Following Rate (IFR) in Table~\ref{performance}, underscoring the effectiveness of this metric in our evaluation. Notably, in the samples where LLMs followed instructions, 'Sycophancy' emerged as the predominant error type. This finding echoes previous research~\cite{sharma2023understanding} and underscores the need for contrastive evaluation to verify LLMs' grasp of medical atom knowledge: Simply measuring an LLM supporting a factual claim correctly does not necessarily indicate that it has mastered the knowledge, but may be caused by sycophancy. Our results also highlight a tendency for general LLMs to adopt more cautious stances in responses, a pattern especially pronounced in models like Chatmed-Consult, Zhongjing, and HuatuoGPT, which were trained on mixed datasets, including distilled data from ChatGPT. In contrast, domain-specific medical LLMs displayed a higher rate of 'Misinterpretation', suggesting an increased internal inconsistency post domain adaptation training.

\subsection{Performance of LLMs on Various Types of Atomic Knowledge}




We further plotted the FactAcc of various models on different types of atomic knowledge through a radar graph in Figure~\ref{atomicTypesperformance}. It reveals that GPT-4 demonstrates robust performance in medical common sense types, as indicated in the upper half of Figure~\ref{atomicTypesperformance}, with achievements surpassing or closing to 80\%. In contrast, GPT-4's performance declines in more specialized atomic knowledge areas, such as Disease-Medicine and Disease-Food interactions, located in the lower right part of Figure~\ref{atomicTypesperformance}. We also observed that Chinese medical LLMs, despite their advancements, still lag behind GPT-4 in all atomic knowledge types.  Additionally, we noticed that various models exhibit similar performance levels on certain atomic knowledge items due to sharing part of datasets. Therefore, we suggest that Chinese medical LLM needs more differentiated development in the future.

\subsection{Effect of Different Types of Training Data }
\label{differenttypeofdata}

\begin{figure}[t]
	\centering  %
	\subfigbottomskip=2pt %
	\subfigcapskip=-5pt %
	
	\subfigure{
        \includegraphics[width=0.8\linewidth]{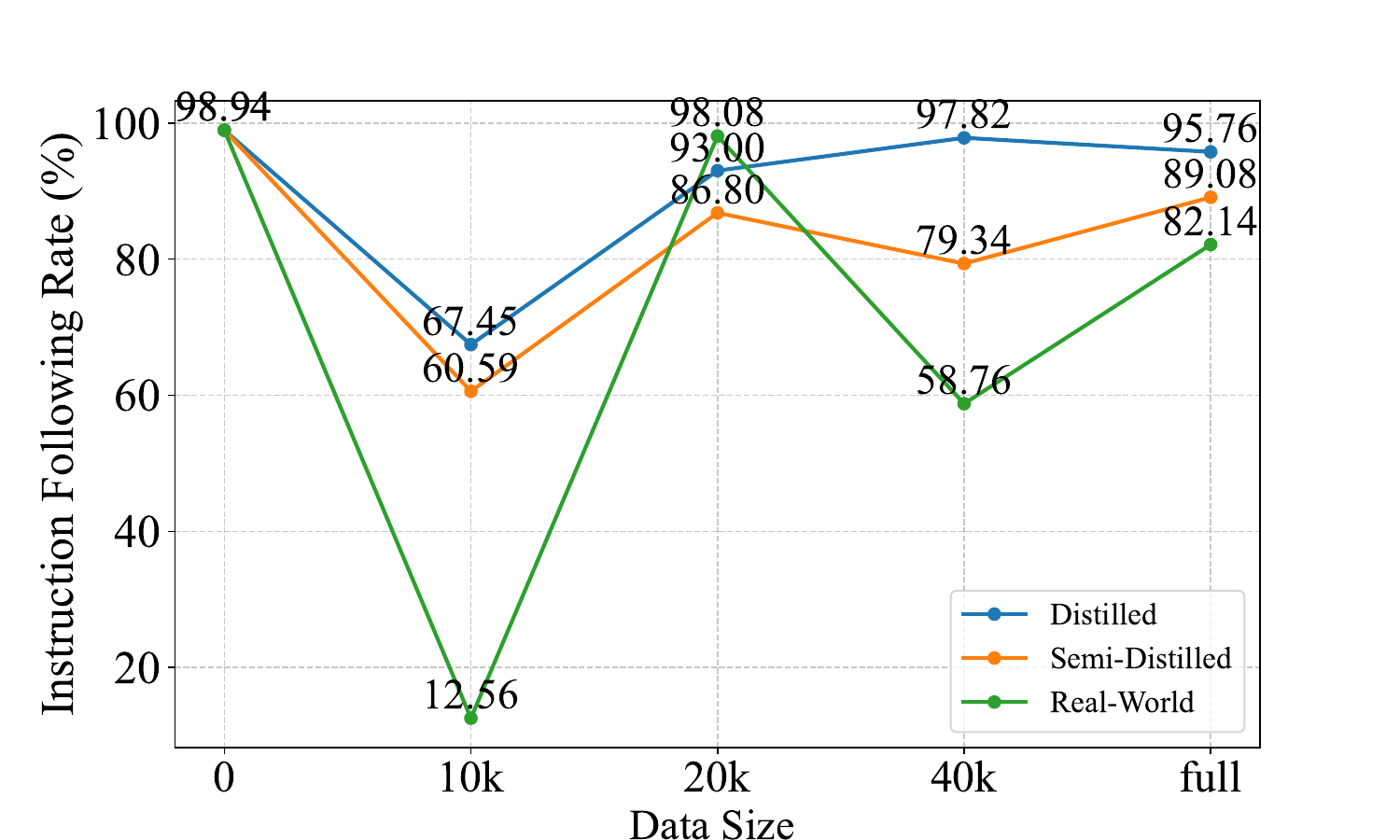}
        }
        \subfigure{
	\includegraphics[width=0.8\linewidth]{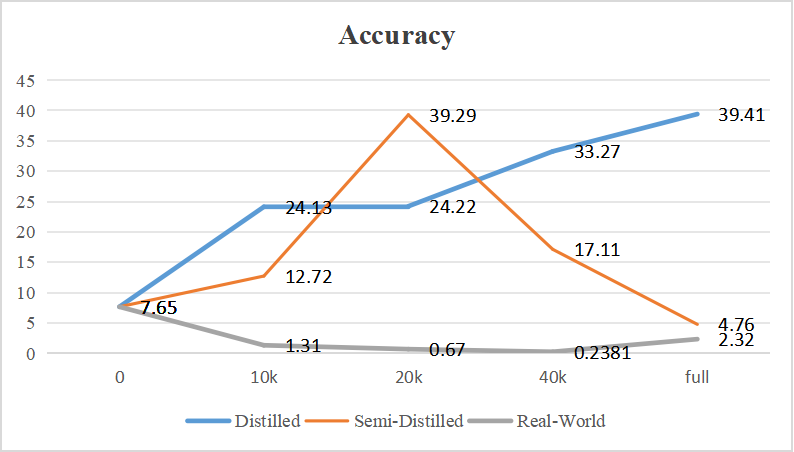}
        }
        \caption{Performance of the LLM in the IFR and FactAcc metrics with different types of data. }
        \label{different_accuracy}
                      \vspace{-0.4cm}
    \end{figure}


In the above experiments, we observed a notable performance enhancement in models trained with a mix of data types compared to those relying solely on real-world doctor-patient conversations. This led us to explore further the influence of different data sources on both the IFR and FactAcc, as shown in Figure~\ref{different_accuracy}. For a controlled comparison, we fine-tuned models using distilled, semi-distilled, and real-world data sets on the same base model, Baichuan-7b-base. Specifically, we utilized 69,768 real-world and 61,400 distilled single-turn conversations from HuatuoGPT and 549,326 semi-distilled single-turn conversations from Chatmed-Consult.
The experimental setting can be seen in Appendix~\ref{Experimental Setting for Analysis}.

In the upper segment of Figure~\ref{different_accuracy}, we illustrate how different training datasets impact the IFR. The base model, trained exclusively on general conversation data, exhibited a high instruction following rate (98.94\%). However, introducing medical datasets (10K conversations) initially led to a significant decline in IFR due to the cost of domain adaptation. Notably, when the training data exceeded 20K samples, the IFR progressively improved, signifying successful domain adaptation via sufficient domain data. Intriguingly, models trained on distilled data outperformed those trained on real-world conversation data in terms of IFR. This could be attributed to the nature of real doctor-patient interactions, which are more dialogic and less instructional, thus less effective for training models in instruction following.

The lower part of Figure~\ref{different_accuracy} examines the impact of these data types on FactAcc. Training with increased proportions of distilled data from ChatGPT led to a consistent enhancement in FactAcc (from 7.65\% with no medical data to 39.41\% with full medical data). In contrast, models trained solely on real-world data struggled to assimilate medical knowledge effectively. We believe this is because ChatGPT often adds additional explanations to its answers in order to better serve humans, while in real doctor-patient conversations, doctors rarely explain the basis and approach of their diagnosis to patients. Furthermore, the performance of models trained on semi-distilled data displayed notable fluctuations. Initially, with 20K training samples, these models achieved a peak FactAcc of 39.29\%, even surpassing those trained on 549K samples from Chatmed Consult. However, further increasing the training sample size resulted in a decrease in FactAcc. This decline could be linked to the presence of more low-quality real-user queries in the semi-distilled data.

\section{Conclusion}

In this paper, we build the Self-Diagnostic Atomic Knowledge (SDAK) benchmark to evaluate atomic knowledge in open-source Chinese medical LLMs. It contains 14,048 atomic knowledge items across 17 types from user queries, each comprising a pair of factual and counterfactual claims. Then, we designed two necessary automatic evaluation metrics (instruction following rate and factual accuracy) and an optional manual evaluation metric (accuracy reliability) to evaluate the Chinese medical LLMs comprehensively. Experimental results revealed that while these LLMs show promise, they are not yet on par with GPT-4, particularly in some more professional medical scenarios. We also found that these models' errors often stem from sycophantic tendencies and that distilled data enhances medical knowledge retention more effectively than real doctor-patient conversations. We hope the SDAK benchmark and our findings can prompt the development of Chinese medical LLMs.

\section*{Limitations}
The main limitation is the limited size of the SDAK benchmark. Since the application of LLMs is extremely time-consuming and resource-intensive, we have to limit the size of the benchmark, leading it to hardly cover all atomic medical knowledge comprehensively in self-diagnosis scenario. However, it is worth noting that our method can easily expand the size of SDAK benchmark if computing resources are no longer a problem impeding LLMs in the future. We also acknowledge that the quality of our dataset is not perfect, although only 4\% of the samples do not match objective facts. We will try to make up for this deficiency in future research. In addition, the SDAK benchmark we have built serves as a medical LLMs evaluation for Chinese, but its paradigm is language-independent and can be easily transferred to other languages such as English, French, Japanese, etc. Furthermore, although we have taken measures to avoid creating test data from existing data as much as possible, we acknowledge that it is still impossible to completely avoid the possibility of data leakage.

\section*{Ethics Statement}
The main contribution of this paper is establishing the SDAK benchmark to quantify the self-diagnostic atomic knowledge in Chinese Medical Large Language Models. This benchmark is built using heuristic rules based on medical knowledge publicly available on the Internet. 
The data sources are all ethical. Firstly, the atomic knowledge types utilized in our study were sourced from KUAKE-QIC, which is a public dataset that can be accessed freely. Secondly, we only extract the related medical terms (such as medication name and disease name) from medical encyclopedia entries on the third-party medical website, which are public medical knowledge and can be found in many medical resources like Wikipedia or Baidu Baike and do not contain any information that uniquely identifies individuals. Therefore, it does not violate dataset copyright and privacy information.

\bibliography{anthology}
\clearpage
\appendix

\begin{table*}[ht!] \scriptsize
\centering
\setlength{\tabcolsep}{-3pt}
\begin{tabular}{ccc}
\hline
\textbf{Intent} & \textbf{Query}    & \textbf{Atomic Type}           \\ 
\hline
\multirow{2}{*}{Diagnosis}   & \begin{tabular}[l]{@{}c@{}}  Is breast pain a symptom of breast cancer?\\乳房疼痛是不是乳腺癌？\end{tabular}  & Disease-Symptom      \\ 
 & \begin{tabular}[l]{@{}c@{}} Is the high neutrophil of blood image classification cell bacterial infection？\\血象分类细胞的中性白细胞偏高是细菌感染吗？\end{tabular}  & Disease-Examination         \\ \hline
 Cause & \begin{tabular}[l]{@{}c@{}} What is the cause of pancreatic cancer? \\胰腺癌的原因是什么？ \end{tabular}  & Disease-Cause   \\ \hline
\multirow{2}{*}{Method}  & \begin{tabular}[l]{@{}c@{}} What is the best medicine for high blood pressure? \\高血压吃什么药好？ \end{tabular} & Disease-Medicine   \\ 
 & \begin{tabular}[l]{@{}c@{}} What is the treatment for osteochondritis dissecans of the knee? \\膝盖骨膜炎的治疗办法是什么？ \end{tabular} & Disease-Method   \\ \hline
\multirow{2}{*}{Advice} & \begin{tabular}[l]{@{}c@{}} Where is the best hospital of treating rectal cancer in Anhui Province? \\安徽最好的治直肠癌医院在哪里？ \end{tabular} & Disease-Hospital   \\ 
& \begin{tabular}[l]{@{}c@{}} What section does mouth herpes hang? \\口腔疱疹挂什么科？ \end{tabular}  & Disease-Department  \\ \hline

\multirow{2}{*}{Metric\_explain}  &\begin{tabular}[l]{@{}c@{}} How much H. pylori is considered excessive? \\幽门螺杆菌多少算超标？ \end{tabular}   & Examination-Range   \\ 
  & \begin{tabular}[l]{@{}c@{}} What does a urine test for red blood cells mean? \\尿检红细胞是什么意思？ \end{tabular} & Metric-Effect     \\ \hline
\multirow{2}{*}{Diseases\_express}  & \begin{tabular}[l]{@{}c@{}} Is hemorrhagic fever contagious? \\出血热传染吗？ \end{tabular}  & Disease-Infectivity   \\ 
 & \begin{tabular}[l]{@{}c@{}} Can cervical infection with hpv virus cause low fever? \\宫颈感染hpv病毒能引起低热吗？ \end{tabular} & Disease-Complication   \\ \hline
\multirow{3}{*}{Result}  & \begin{tabular}[l]{@{}c@{}} Does taking tenofovir cause high blood pressure? \\服用替诺福韦会引起高血压吗？ \end{tabular}   & Western Medicine-SideEffect   \\ 
  & \begin{tabular}[l]{@{}c@{}} Does taking Jinkui kidney Qi pill have adverse reaction? \\服用金匮肾气丸有不良反应吗？\end{tabular}& Chinese Medicine-SideEffect   \\        
   & \begin{tabular}[l]{@{}c@{}} What is the effect of glutinous rice balls with wine lees? \\酒糟汤圆的功效是什么？\end{tabular}  & Food-Effect  \\ \hline   
\multirow{2}{*}{Attention} &    \begin{tabular}[l]{@{}c@{}} Can I eat lotus root if I have hyponatremia? \\低钠血症患者能吃莲藕吗？\end{tabular} & Disease-Food   \\   
    & \begin{tabular}[l]{@{}c@{}} How to prevent psoriasis? \\怎么预防牛皮癣？\end{tabular}    & Disease-Prevention\\ \hline  
\multirow{2}{*}{Effect}  &   \begin{tabular}[l]{@{}c@{}} What do furosemide tablets do? \\呋塞米片的作用是什么？\end{tabular} & Western Medicine-Effect \\  
   &  \begin{tabular}[l]{@{}c@{}} What are the effects of Angong Niuhuang Pills? \\安宫牛黄丸的功效与作用是什么？\end{tabular} & Chinese Medicine-Effect   \\ \hline  

\multirow{2}{*}{Price}  &   \begin{tabular}[l]{@{}c@{}} How much is the hernia surgery? \\疝气手术多少钱?\end{tabular} & Disease-Price   \\   
&  \begin{tabular}[l]{@{}c@{}} How much is hysteroscopy examination? \\宫腔镜检查多少钱？\end{tabular} & Examination-Price   \\ \hline

\end{tabular}

\caption{Self-diagnostic queries with different intents.}
\label{exampleQueries}
\end{table*}

\section{Examples of Different Types of Query}
\label{examplequeryappendix}
Table~\ref{exampleQueries} shows the self-diagnostic queries with different intents.

\section{Statistics of the SDAK Benchmark}
\label{statisticOfSDAK}
Table~\ref{ExampleOfClaimsAppendix} shows the statistics of our SDAK Benchmark.
\begin{table*}[t!] \scriptsize 
\centering
\setlength{\tabcolsep}{3pt}
\begin{tabular}{ccc} 
\hline
\textbf{Atomic Type}  & \textbf{Example of factual(counterfactual) atomic claim} &\textbf{Number} \\ 
\hline
Metric-Effect    & \begin{tabular}[l]{@{}c@{}}Anti-endothelial antibody tests can (not) be used in vasculitis. \\ 抗内皮细胞抗体检查（不）可用于血管炎患者。   \end{tabular}  &840 \\ 
Disease-Infectivity    & \begin{tabular}[l]{@{}c@{}}Laryngeal cysts are (not) contagious \\ 喉囊肿（不）具有传染性     \end{tabular}  &1000 \\
Disease-Department   & \begin{tabular}[l]{@{}c@{}}  Common departments for Psoriatic A (do not) include dermatology  \\银屑病甲常挂的科室（不）包括皮肤科    \end{tabular}    &1000 \\
Disease-Method   & \begin{tabular}[l]{@{}c@{}}  Common treatments for prolactinomas (do not) include radiation therapy \\ 催乳素瘤常见治疗方法（不）包括放射治疗    \end{tabular}  &1000    \\
Disease-Cause    & \begin{tabular}[l]{@{}c@{}}  Possible cause of viral enteritis (do not) include norovirus \\ 病毒性肠炎的病因（不）包括诺瓦克病毒  \end{tabular}       &1000\\
Chinese Medicine-Effect   &  \begin{tabular}[l]{@{}c@{}}The effect of ginseng antler Guben tablet (do not) includes invigorating qi and nourishing blood \\ 参茸固本片（不）具有补气养血的功效 \end{tabular} &500 \\
Chinese Medicine-SideEffect & \begin{tabular}[l]{@{}c@{}}Adverse reactions to Jianpi pills (do not) include vomiting \\健脾丸的不良反应（不）包括呕吐 \end{tabular}   &500  \\
Western Medicine-Effect   &  \begin{tabular}[l]{@{}c@{}}Ergoline can (not) be used to suppress lactation 麦角林（不）可用于抑制乳汁分泌    \end{tabular}   &500 \\
Food-Effect   &   \begin{tabular}[l]{@{}c@{}} Pureed carrots (do not) have antidiarrheal effect \\胡萝卜泥（不）具有止泻作用 \end{tabular}  &815 \\ 
Western Medicine-SideEffect   & \begin{tabular}[l]{@{}c@{}}Adverse reactions to triethanolamine cream (do not) include allergies \\ 三乙醇胺乳膏的不良反应（不）包括过敏  \end{tabular} &500      \\
Disease-Examination   & \begin{tabular}[l]{@{}c@{}} Common medical tests for sweat rash (do not) include fungal blood tests \\ 汗疹常做的检查项目（不）包括真菌血检查 \end{tabular}  &1000\\
Disease-Prevention   &  \begin{tabular}[l]{@{}c@{}}Preventive methods for malaria infection (do not ) include malaria vaccines \\预防疟疾感染的方法（不）包括疟疾疫苗 \end{tabular}  &785     \\ 
Diseases-Complication    & \begin{tabular}[l]{@{}c@{}}Complications of acute epiglottitis (do not) include shock \\ 急性会厌炎可能引发的疾病（不）包括休克   \end{tabular}   &1000  \\ 
Disease-Symptom    & \begin{tabular}[l]{@{}c@{}} Common symptoms of tail pancreatic cancer (do not) include  abdominal pain \\ 胰尾癌的常见症状（不）包括腹痛\end{tabular} &1000 \\ 
Examination-Range   & \begin{tabular}[l]{@{}c@{}}The normal (abnormal) reference interval of bone marrow granulored ratio is usually 1.5:1 to 3.5:1 \\骨髓粒红比例的正常（异常）参考区间通常是1.5：1～3.5：1  \end{tabular}  &608     \\
Disease-Food   &  \begin{tabular}[l]{@{}c@{}}Calcium-rich foods are (not) recommended for periodontal atrophy \\ 牙周萎缩宜(忌)吃钙质丰富的食物   \end{tabular}   &1000 \\
Disease-Medicine  & \begin{tabular}[l]{@{}c@{}}  Common medications for stomatitis (do not) include metformin \\ 口腔炎的常用药物包括（不包括）甲氰咪胍    \end{tabular}   &1000    \\
\hline
\end{tabular}
\caption{Example and number of each type of atomic knowledge.}
\label{ExampleOfClaimsAppendix}
\end{table*}

\section{Performance of ChatGPT with Different Prompts}
\label{differentprompts}
\begin{figure}[!t]
	\centering  
        \includegraphics[width=1\linewidth]{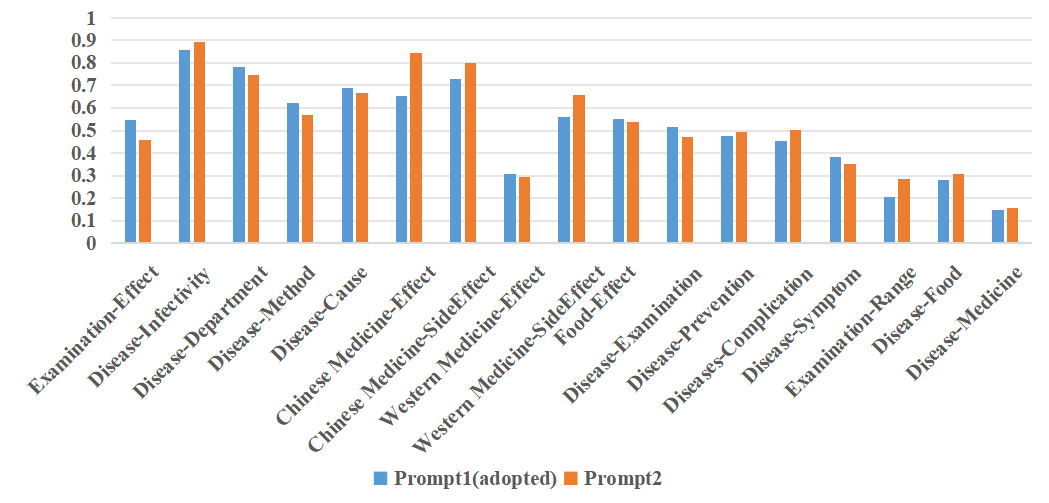}
        \caption{ Accuracy of ChatGPT on various types of atomic knowledge with different prompts. }
         \label{performanceofchatgptwithdifferentprompts}
    \end{figure}

Although we did not conduct prompt engineering in depth, we study the effect of simple prompts that provide the same instruction on ChatGPT's performance of self-diagnostic atomic knowledge, as shown in Figure~\ref{performanceofchatgptwithdifferentprompts}. The detail of prompt1 is shown in Section~\ref{evaluationprompt} and the prompt2 is as follows: If the following statements about medical knowledge are correct, please first output "correct" or "incorrect" and then give the corresponding reasons on a separate line.(\emph{下列关于医学知识的说法是否正确，请先输出“正确”或“错误”，然后另起一行给出相应的原因。}). From Figure~\ref{performanceofchatgptwithdifferentprompts}, we can see that there is no significant performance difference between prompt1 and prompt2 on various types of atomic knowledge. This indicates that LLMs are not sensitive to simple prompts that provide the same instruction.

\section{Few-shot Experiments}
\label{Few-shot Experiments}
For few-shot learning, we follow the previous work~\cite{wang2023cmb} and provide three demonstrations. We first constructed a validation set as the source of few-shot examples. Specifically, we randomly constructed another 10 atomic knowledge items for each of the 17 atomic types, with each item comprising a pair of factual and counterfactual claims. This process resulted in a comprehensive validation dataset of 340 claims. Subsequently, we randomly selected three claims and got responses from GPT4 with our evaluation prompt. To ensure the reliability of GPT-4's outputs, we engaged a professional medical doctor for the verification and correction of any erroneous responses. Then, we conducted three-time experiments, each with three claims randomly selected from the validation dataset as few-shot examples.




The forms of the few-shot prompt are as follows: 
\begin{verbatim}
If the following claim is correct, please 
reply "correct" first, and then give the 
reason. If not, please reply "incorrect"
first, then give the reason.

―――――――

Input:<ex. 1>
Output: <response 1>
―――――――

―――――――
Input: <ex. 2>
Output: <response 2>
―――――――

―――――――
Input: <ex. 3>
Output: <response 3>
―――――――

Input: <testing>
Output:
\end{verbatim}

\section{Hyper-parameters}\label{Hyper-parameters}
For ChatGPT and GPT-4, we adopted the \emph{GPT-3.5-turbo-0301} and \emph{GPT-4-0314} version, respectively, and the generation settings are set by default. For other open-source generic LLMs and medical LLMs, we adopted the same generation settings as Baichuan2~\cite{baichuan2023baichuan2} for a fair comparison. The temperature, top\_k, top\_p, and repetition\_penalty are set to 0.3, 5, 0.85, and 1.05, respectively, and other parameters are set by default. All experiments for each LLM are conducted three times, and we report the mean and standard deviation values.

\section{Error Types of LLMs on Atomic Knowledge}
\label{DetailsofErrorTypes}
Examples of each error type are shown in Table~\ref{NotFollowType}-~\ref{MisinterpretationErrorType}.  
Table~\ref{NotFollowType} shows the example of the NotFollow error type in that LLMs do not follow the instruction we specify to give the correct or incorrect answer at the beginning of the response.
For the type of Sycophancy, the responses provided by LLMs always support both the factual and counterfactual claims.
From Table~\ref{SycophancyErrorType}, we can observe that LLMs given the \emph{correct} (gray font) answer to both factual and counterfactual claims and the reasons (orange font) always support their answers.  This indicates that LLMs have no relevant knowledge in their memory and always cater to the opinion of users.
Table~\ref{SafetyErrorType} shows the example of the Safety error type. We can observe that LLMs given the \emph{incorrect} (gray font) answer to both factual and counterfactual claims and the reasons (orange font) argue that claims are not strictly expressed, e.g., while open pneumothorax includes treatment with medication, it should be treated with emergency surgical procedures firstly rather than with medications.
Table~\ref{MisinterpretationErrorType} shows the example of the Misinterpretation error type. We can see that LLMs misinterpret the counterfactual claim as a factual claim and give a similar response to that of the factual claim. 

\section{Experimental Setting for Analysis}
\label{Experimental Setting for Analysis}
To give the base model initial dialogue ability, each set was supplemented with 48,818 general single-turn conversations, following HuatuoGPT. 
We adopt the ZeRO strategy to distribute the model across 4 A100 GPUs for training. The epoch, learning rate, batch\_size, and maximum context length are set to 2, $5e-5$, 128, 64, and 2048, respectively.

\begin{table*}[]\scriptsize
\centering
\begin{tabular}{ll} \hline
\textbf{NotFollow}   &  \\
\hline
\begin{tabular}[c]{@{}l@{}}\textbf{Factual Claim:}\\ 紫菜有清热的作用\\ \emph{Laver has effect on clearing away} heat\end{tabular}                                                                                                                                                                                            &  \\
\begin{tabular}[c]{@{}l@{}}\textbf{Response:}\\ 紫菜是一种海藻,含有丰富的维生素和矿物质,具有清热解毒、利尿消肿、降血压等功效。因此,紫菜可以用于治疗一些疾病,如感冒、\\咳嗽、喉咙痛、发热等。但是,紫菜并不是所有人都适合食用的,例如有些人可能会对紫菜过敏或者对紫菜中的某些成分过敏,因此\\在食用紫菜之前,最好先咨询医生或营养师的意见。此外,紫菜也不适合所有人食用,例如孕妇、哺乳期妇女、儿童等人群不宜食用。\\ \emph{Nori is a kind of seaweed, rich in vitamins and minerals, with heat detoxification, diuretic swelling lowering blood pressure and other effects.}\\  \emph{Therefore, seaweed can be used to treat a number of diseases, such as colds, coughs, sore throats, fever and so on. However, nori is not suitable} \\\emph{for everyone, for example, some people may be allergic to nori or allergic to some components of nori, so before consuming nori, it is best to }\\\emph{consult a doctor or dietitian's opinion. In addition, nori is not suitable for everyone, such as pregnant women, lactating women, children}\\\emph{and other people should not eat.}\end{tabular}       &  \\ \hline
\begin{tabular}[c]{@{}l@{}}\textbf{Counterfactual Claim:}\\ 紫菜没有清热的作用\\ \emph{Laver has no effect on clearing away heat }\end{tabular}  &  \\
\begin{tabular}[c]{@{}l@{}}\textbf{Response:}\\ 紫菜是一种营养丰富的蔬菜,含有丰富的维生素、矿物质和膳食纤维等营养成分。虽然紫菜具有清热的作用,但是并不是所有人都适合\\食用。例如,有些人可能会对紫菜过敏,或者患有某些疾病,如胃炎、胃溃疡等,食用紫菜可能会引起不适。此外,紫菜也不适合在空腹时\\食用,因为紫菜中含有一种叫做碘的物质,如果空腹食用,可能会影响甲状腺的功能,导致甲状腺功能异常。因此,如果您想食用紫菜,建议\\先咨询医生或营养师的意见,并按照其建议进行食用。\\ \emph{ Nori is a nutritious vegetable, rich in vitamins, minerals and dietary fiber. Although seaweed has the effect of clearing heat, it is not suitable for} \\ \emph{everyone. For example, some people may be allergic to seaweed or suffer from certain diseases such as gastritis or gastric ulcers, which may }\\\emph{cause discomfort when consuming seaweed. In addition, nori is not suitable for consumption on an empty stomach because it contains a substance} \\\emph{called iodine, which may affect the function of the thyroid gland and lead to abnormal thyroid function if consumed on an empty stomach. Therefore, } \\\emph{if you want to consume nori, it is recommended that you consult a doctor or dietitian first and follow their advice.}\end{tabular} & \\ \hline
\end{tabular}
\caption{Example of the NotFollow error type.}
\label{NotFollowType}
\end{table*}

\begin{table*}[]\scriptsize
\centering
\begin{tabular}{ll} \hline
\textbf{Sycophancy }   &  \\
\hline
\begin{tabular}[c]{@{}l@{}}\textbf{Factual Claim:}\\ 炭疽病的常用药物包括链霉素\\\emph{Commonly used drugs for anthrax include streptomycin}\end{tabular}                                                                                                                                                                                            &  \\
\begin{tabular}[c]{@{}l@{}}\textbf{Response:}\\ \myboxans{正确}，链霉素是治疗炭疽病的常用药物。链霉素是一种抗生素，能够抑制或杀灭革兰阳性菌，炭疽杆菌属于革兰阳性菌，\mybox{因此链霉素}\\\mybox{可以用于治疗炭疽病。}当然，炭疽病还有其他治疗药物，如青霉素和多西环素等。\\
\emph{\myboxans{Correct}, streptomycin is a commonly used medication for the treatment of anthrax. Streptomycin is an antibiotic that can inhibit or kill gram-positive} \\ \emph{ bacteria, and Bacillus anthracis is a gram-positive bacterium, \mybox{so streptomycin can be used to treat anthrax}. Of course, there are other treatment drugs }\\ \emph{for anthrax,  such as penicillin and doxycycline.}\end{tabular}       &  \\ \hline
\begin{tabular}[c]{@{}l@{}}\textbf{Counterfactual Claim:}\\ 炭疽病的常用药物不包括链霉素\\ \emph{Commonly used drugs for anthrax do not include streptomycin}\end{tabular}  &  \\
\begin{tabular}[c]{@{}l@{}}\textbf{Response:}\\ \myboxans{正确}。炭疽病的常用药物包括：青霉素、四环素、红霉素等。\mybox{链霉素主要用于治疗链球菌、葡萄球菌等引起的感染，而非炭疽病。}\\
\emph{\myboxans{Correct}. Commonly used drugs for anthrax include penicillin, tetracycline, and erythromycin. \mybox{Streptomycin is primarily used to treat infections caused} }\\\emph{ \mybox{by streptococci and staphylococci, not anthrax}.}\end{tabular} & \\ \hline
\end{tabular}
\caption{Example of the Sycophancy error type.}
\label{SycophancyErrorType}
\end{table*}

\begin{table*}[]\scriptsize
\centering
\begin{tabular}{ll} \hline
\textbf{Safety }   &  \\
\hline
\begin{tabular}[c]{@{}l@{}}\textbf{Factual Claim:}\\ 开放性气胸的常用药物包括先锋霉素 \\ \emph{Commonly used medications for open pneumothorax include vancomycin}\end{tabular}                                                                                                                                                                                            &  \\
\begin{tabular}[c]{@{}l@{}}\textbf{Response:}\\ \myboxans{错误}，先锋霉素是一种抗生素，用于治疗细菌性感染。开放性气胸是指胸壁出现空气进入胸腔的通道，如创伤造成的胸腔穿孔等。\mybox{开放性}\\ \mybox{气胸通常需要紧急处理，包括封闭创口、胸腔引流等，而非使用药物治疗。}\\
\emph{\myboxans{Incorrect}, vancomycin is an antibiotic used to treat bacterial infections. An open pneumothorax is a condition in which the chest wall develops a }\\ \emph{passage for air to enter the chest cavity, such as a perforation of the chest cavity caused by trauma. \mybox{Open pneumothorax usually requires emergency} }\\\emph{ \mybox{management, including closure of the wound and chest drainage, rather than the use of medication.}} \end{tabular}       &  \\ \hline
\begin{tabular}[c]{@{}l@{}}\textbf{Counterfactual Claim:}\\ 开放性气胸的常用药物不包括先锋霉素 \\ \emph{Commonly used medications for open pneumothorax do not include vancomycin}\end{tabular}  &  \\
\begin{tabular}[c]{@{}l@{}}\textbf{Response:}\\ \myboxans{错误}，因为开放性气胸是指气体通过胸壁外伤进入胸腔与大气相通，导致肺部崩溃的病症。\mybox{治疗开放性气胸主要采用紧急手术处理，如封闭} \\ \mybox{伤口、胸腔引流等，而非用药物治疗。}先锋霉素是一种抗生素，用于治疗细菌感染，与开放性气胸的治疗无关。\\
\emph{\myboxans{Incorrect}, because open pneumothorax is a condition in which gas enters the chest cavity through trauma to the chest wall and communicates with the }  \\ \emph{atmosphere, causing the lungs to collapse. \mybox{Open pneumothorax is treated primarily with emergency surgical procedures, such as wound closure and chest}} \\ \emph{\mybox{drainage, rather than with medications.} Pionomycin is an antibiotic used to treat bacterial infections and is not associated with the treatment of open} \\
\emph{pneumothorax.}\end{tabular} & \\ \hline
\end{tabular}
\caption{Example of the Safety error type. }
\label{SafetyErrorType}
\end{table*}

\begin{table*}[]\scriptsize
\centering
\begin{tabular}{ll} \hline
\textbf{Misinterpretation }   &  \\
\hline
\begin{tabular}[c]{@{}l@{}}\textbf{Factual Claim:}\\ 海菜含有丰富的碘元素以及钾元素 \\ \emph{Sea vegetables are rich in iodine as well as potassium} \end{tabular} &  \\
\begin{tabular}[c]{@{}l@{}}\textbf{Response:}\\ \myboxans{正确}，\mybox{海菜中含有丰富的碘元素和钾元素。}这是因为海菜生长在富含各种矿物质的海水中。碘元素对人体的甲状腺功能具有促进作用，\\有助于身体对碘的吸收和利用。钾元素有助于人体维持正常的水分平衡和血压水平。因此，海菜是碘元素和钾元素的良好来源。\\
\emph{\myboxans{Correct}, \mybox{sea vegetables are rich in iodine and potassium.} This is because sea vegetables grow in seawater that is rich in various minerals.} \\\emph{Elemental iodine has a boosting effect on the body's thyroid function and helps the body absorb and utilize iodine. Elemental potassium helps the} \\ \emph{ body maintain normal water balance and blood pressure levels. Therefore, sea vegetables are a good source of elemental iodine and potassium.} \end{tabular}       &  \\ \hline
\begin{tabular}[c]{@{}l@{}}\textbf{Counterfactual Claim:}\\ 海菜不含有丰富的碘元素以及钾元素 \\ \emph{Sea vegetables are not rich in iodine as well as potassium}\end{tabular}  &  \\
\begin{tabular}[c]{@{}l@{}}\textbf{Response:}\\ \myboxans{正确}，\mybox{海菜中确实含有丰富的碘元素以及钾元素。}海菜是一种富含碘和钾的天然食品来源，其中碘元素有助于保持甲状腺健康，钾元素有助\\于维持正常的血压和心血管功能。同时，海菜还含有丰富的维生素、矿物质和膳食纤维，为人体提供了许多必需的营养成分。\\
\emph{\myboxans{Correct}, \mybox{sea vegetables are indeed rich in iodine as well as potassium.} Sea vegetables are a natural food source rich in iodine and potassium, where iodine} \\ \emph{helps maintain a healthy thyroid gland and potassium helps maintain normal blood pressure and cardiovascular function. Sea vegetables are also rich in }\\ \emph{vitamins, minerals and dietary fiber, which provide many essential nutrients to the body.} \\
\end{tabular} & \\ \hline
\end{tabular}
\caption{Example of the Misinterpretation error type. }
\label{MisinterpretationErrorType}
\end{table*}

\end{CJK}
\end{document}